%% file: main.tex
\documentclass[acmtog]{acmart}

\usepackage{multirow}
\usepackage{booktabs}
\usepackage{enumitem}

\copyrightyear{2026}
\acmYear{2026}
\setcopyright{cc}
\setcctype{by-nc-nd}
\acmConference[SA Conference Papers '26]{SIGGRAPH Asia 2026 Conference Papers}{December 01--04, 2026}{Kuala Lumpur, Malaysia}
\acmBooktitle{SIGGRAPH Asia 2026 Conference Papers (SA Conference Papers '26), December 01--04, 2026, Kuala Lumpur, Malaysia}
\acmDOI{10.1145/3829340.3842252}
\acmISBN{979-8-4007-2842-6/2026/12}

\usepackage[ruled]{algorithm2e} %

\SetAlFnt{\small}
\SetAlCapFnt{\small}
\SetAlCapNameFnt{\small}
\SetAlCapHSkip{0pt}

\newcommand{\best}[1]{\underline{#1}}
\newcommand{\bestbold}[1]{\textbf{#1}} %

\begin{document}
\title{JointLight: Human-Centric Video Relighting with Asynchronous RGB-Irradiance Joint Modeling}

\author{Henglei Lv}
\orcid{0009-0008-1907-0996}
\affiliation{%
 \institution{Institute of Computing Technology, Chinese Academy of Sciences, China and University of Chinese Academy of Sciences, China
 }
 \streetaddress{6 Kexueyuan Nanlu}
 \city{Haidian Qu}
 \state{Beijing}
 \postcode{100190}
 \country{China}
}
\email{lvhenglei22s@ict.ac.cn}

\author{Bailin Deng}
\affiliation{%
 \institution{Cardiff University}
 \city{Cardiff}
 \country{United Kingdom}
}
\email{DengB3@cardiff.ac.uk}

\author{Xiaoqiang Liu}
\affiliation{%
\institution{Kuaishou Technology}
 \city{Haidian Qu}
 \state{Beijing}
\country{China}}
\email{liuxiaoqiang@kuaishou.com}

\author{Yichen Wang}
\affiliation{%
 \institution{Institute of Computing Technology, Chinese Academy of Sciences, China and University of Chinese Academy of Sciences, China}
 \city{Haidian Qu}
 \state{Beijing}
 \country{China}
}
\email{wangyichen191@mails.ucas.ac.cn}

\author{Haoxian Zhang}
\affiliation{%
 \institution{Kuaishou Technology}
 \city{Haidian Qu}
 \state{Beijing}
 \country{China}
}
\email{zhanghaoxian@kuaishou.com}
\author{Pengfei Wan}
\affiliation{%
 \institution{Kuaishou Technology}
 \city{Haidian Qu}
 \state{Beijing}
 \country{China}
 }
\email{wanpengfei@kuaishou.com}

\author{Lin Gao}
\authornote{Corresponding author is Lin Gao (gaolin@ict.ac.cn).}
\affiliation{%
 \institution{Institute of Computing Technology, Chinese Academy of Sciences, China and University of Chinese Academy of Sciences, China}
 \streetaddress{6 Kexueyuan Nanlu}
 \city{Haidian Qu}
 \state{Beijing}
 \postcode{100190}
 \country{China}
}
\email{gaolin@ict.ac.cn}

\renewcommand\shortauthors{Lv, H. et al}

\begin{abstract}
Understanding and modeling light transport in human-centric scenes is a longstanding challenge. While recent progress in diffusion models has boosted research in relighting, precise illumination control still lags, primarily due to the scarcity of diverse, high-quality, precisely labeled video relighting datasets. Existing end-to-end relighting approaches model light transport implicitly through neural networks, which substantially complicates controllability. In contrast, some works resort to path-traced irradiance cues as an explicit blueprint for light transport. However, such constructed irradiance cues are prone to artifacts and hinder generative quality when directly used as control signals.
With the key observation that noise corruption removes high-frequency artifacts and preserves low-frequency light transport, we introduce an asynchronous joint-modeling paradigm that facilitates fine-grained illumination control. During two-stage training, we force the model to jointly denoise RGB and ground-truth irradiance frames with separate timesteps scheduling. A domain matching strategy is applied to the second stage to bridge the domain gap between irradiance cues and ground-truth irradiance maps. At inference time, we introduce an asynchronous hijacked generation pipeline that facilitates flexible illumination control. For training data, we develop a multi-illumination high-quality synthetic video dataset with paired irradiance maps using a powerful game engine. Trained exclusively on this dataset, our model exhibits robust generalization to real-world scenarios and outperforms prior approaches in both lighting fidelity and generative quality. Project page: \url{http://www.geometrylearning.com/JointLight/}

\end{abstract}



\begin{teaserfigure}
  \centering  \includegraphics[width=\textwidth]{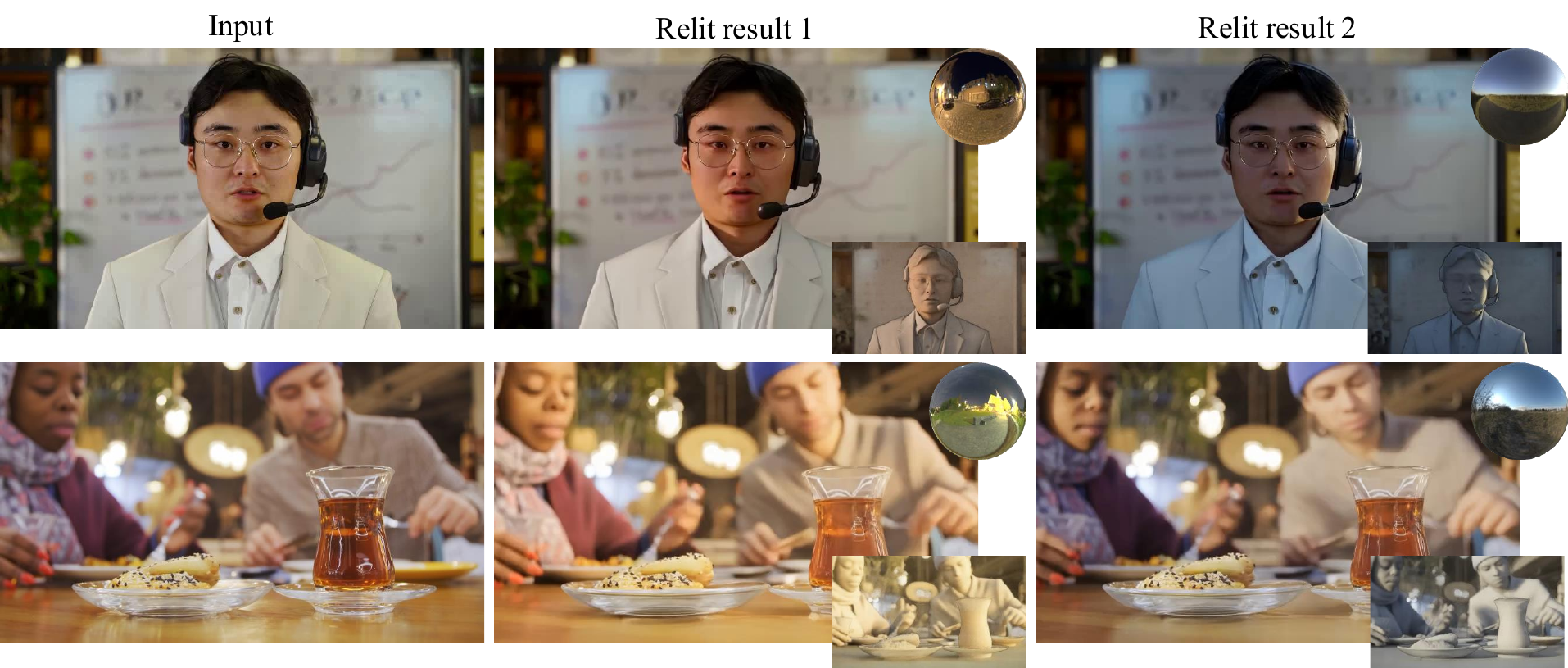}
  \caption{Our method jointly models RGB and irradiance map, enabling high-quality generation with physically accurate illumination control.
  }
  \Description{This is the teaser figure for the article. }
  \label{fig:teaser}
\end{teaserfigure}

\maketitle

\input{body}

\end{document}

%% file: body.tex
\section{Introduction}

Lighting fundamentally affects how we perceive a scene. Modifying lighting while preserving scene identity enables a wide range of applications, such as artistic creation, film production, augmented reality, and computer vision research. Fundamentally, relighting requires an understanding of how light interacts with the scene. Traditional graphics pipelines simulate physical light transport with explicit geometry and materials. For objects and portraits, meticulously calibrated studio equipment~\cite{debevec2012light} is used to reconstruct geometry and capture reflectance fields. However, real-world scenes comprise complex combinations of objects and persons, of which the geometry and material properties are impractical to capture. Relighting of human-centric scenes in diverse indoor and outdoor environments thus remains a challenging task.

With the advent of powerful generative models, relighting with a single image or video input has drawn increasing attention. It is an innately ill-posed problem, since geometry, material, and lighting are all coupled in the single input. Some works~\cite{zhang2025scaling,fang2025relightvid,liu2025unilumos} construct degradation pairs for real-world images or videos and train end-to-end diffusion models for relighting. These methods, lacking explicit physical light transport modeling, are typically constrained to text prompts or environment maps as conditioning inputs, failing to offer precise control over lighting parameters such as light source angles. Some other works~\cite{liang2025diffusion,he2025unirelight} utilize graphics engines to construct synthetic data, enabling relighting with environment maps as a condition. Such synthetic data are limited to simple object-centric scenes, hindering generalization to complex human-centric real-world scenarios. More importantly, implicit light transport modeling within neural networks cannot guarantee physical accuracy.

\begin{figure*}[ht]
    \centering
    \includegraphics[width=\linewidth]{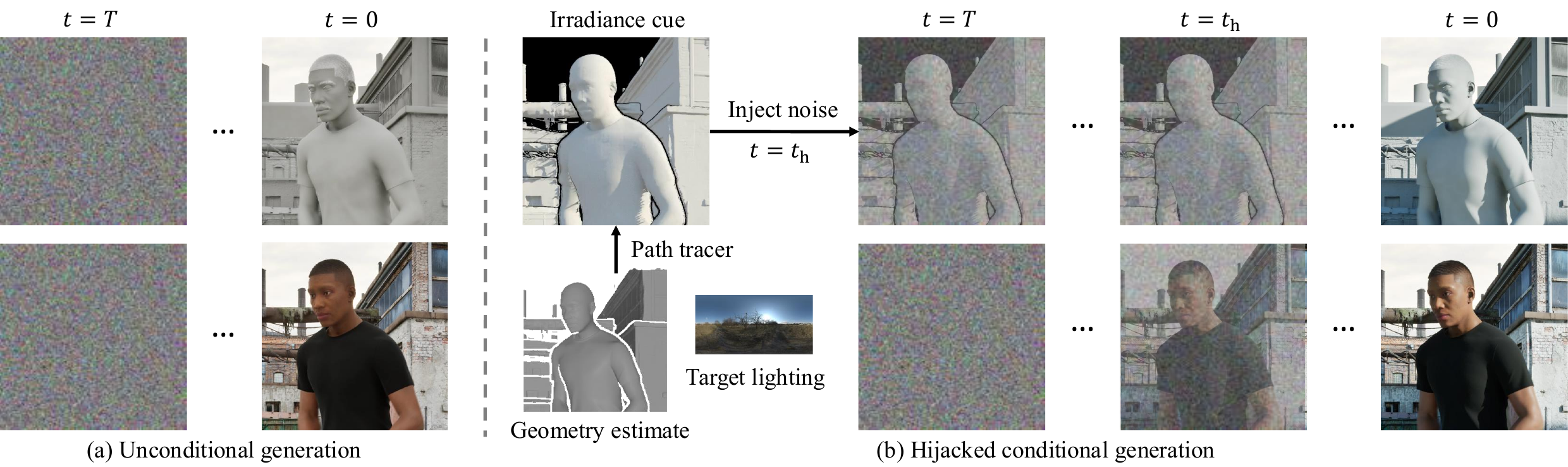}
    \caption{
    Inference pipeline. From the input video and target environment map, we render a per-frame irradiance cue using monocular point-cloud estimation and Mitsuba 3. The cue is noised to $t_h$, kept fixed until $t_h$, then jointly denoised with RGB, preserving low-frequency transport while suppressing artifacts.}
    \label{fig::infer}

\end{figure*}

Recent advances in monocular geometry estimation~\cite{wang2025moge,wang2025moge2,lin2025depth,mescheder2025sharp,wang2025vggt} have opened new avenues for scene relighting.
While monocular point clouds serve as a coarse geometric proxy, they are unsuitable for direct integration into traditional graphics pipelines due to single-layer geometry and artifacts from depth discontinuities at object boundaries.
With a diffuse white BRDF and physics-based rendering, this proxy produces an irradiance cue that characterizes the scene's diffuse light transport, as illustrated in Figure~\ref{fig::infer}.  Although this irradiance cue remains plagued by artifacts and exhibits a significant domain gap relative to the ground-truth irradiance map, we propose that this discrepancy can be effectively bridged through noise corruption. The key observation is that after blending with a certain amount of Gaussian noise, the rough irradiance cue and ground-truth irradiance map become indistinguishable.
In this way, noise corruption mitigates the domain gap by removing high-frequency artifacts and preserving low-frequency content. We then model this rough irradiance distribution via the diffusion process. During training, the model jointly denoises RGB video and irradiance maps with separate timestep scheduling, learning to transform a noised irradiance cue into a pixel-aligned fine-grained irradiance map. At inference time, we render the irradiance cue using a geometry estimate and the target lighting with a path tracer. The irradiance cue is then noised and jointly denoised with the RGB video.

For training data, we develop a large-scale human-centric multi-illumination video dataset with paired irradiance maps using a powerful game engine. It contains 68K synthetic videos from 6.8K dynamic scenes across 30 artist-made 3D environments. A character performing random actions is placed in the center of the camera, where random camera shake and camera movement are applied. Each scene is rendered with 10 different lighting conditions sampled from 821 HDR environment maps, of which 20\% are set to rotating.

Experimental results show that our method outperforms state-of-the-art realistic video relighting approaches in both relighting fidelity and video quality. Ablation studies verify the effectiveness of our key designs.
To summarize, we present three main contributions:
\begin{itemize}[leftmargin=*,topsep=1pt]
\item An asynchronous RGB-irradiance joint-modeling approach with separate timestep scheduling, plus a domain matching strategy that bridges the gap between noisy irradiance cues and ground-truth irradiance maps.
\item A high-quality human-centric multi-illumination video dataset with paired irradiance maps across diverse indoor and outdoor lighting conditions. This dataset will be publicly released to advance research in human-centric video relighting and intrinsic decomposition.
\item Extensive experiments show that our method achieves state-of-the-art results on video relighting in both visual quality and lighting fidelity.
\end{itemize}

\begin{figure*}[ht]
    \centering
    \includegraphics[width=\linewidth]{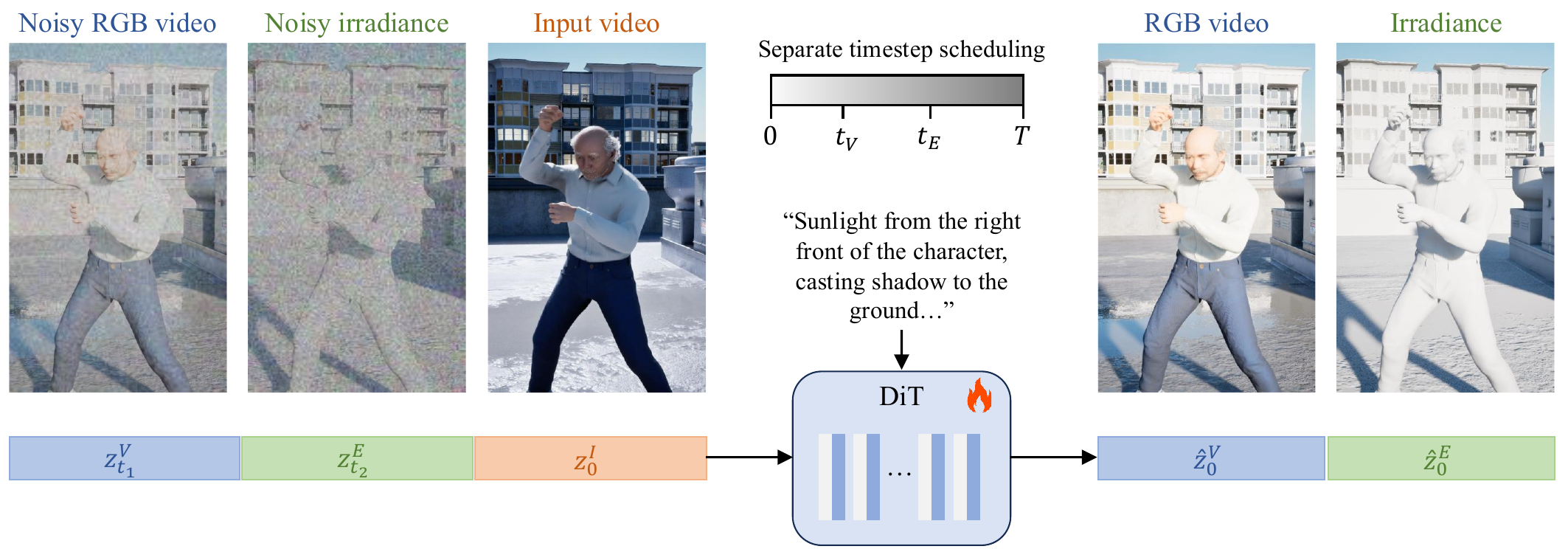}
    \caption{ Overall training pipeline. Noisy RGB video, noisy irradiance video and input video are encoded and concatenated frame-wise before fed into the fully tuned Diffusion Transformer. Noisy RGB and irradiance tokens are assigned with separate noise timesteps to enable asynchronous inference (Section~\ref{sec::inference}).
   }
    \label{fig::pipeline}
\end{figure*}

\section{Related Work}
\paragraph{Human-centric scene relighting.}
While object-centric~\cite{jin2024neural,zeng2024dilightnet,wu2024deferredgs,bharadwaj2025genlit,mao2025neuspir} and portrait-centric~\cite{lv2025gsheadrelight,jiang2023nerffacelighting,saito2024relightable,chaturvedi2025synthlight,mei2025lux,wang2025comprehensive,zhang2024physmatlight} static relighting have been extensively explored, human-centric video relighting in dynamic indoor and outdoor scenes remains underexplored.
Many works~\cite{zhang2025scaling,fang2025relightvid,liu2025unilumos,liu2025light,zhou2025light,zeng2025lumen,liu2026dreamlight,lin2026illumicraft,zhao2024sgformer} rely on degradation pairs with text prompts or environment maps, but provide only coarse control and often produce stylized results. \citep{liang2026pi} use path tracing rendering as an anchor point to online optimize static relighting results, which cannot extend to dynamic scenes.
DiffusionRenderer~\cite{liang2025diffusion} curates synthetic multi-illumination data with simple objects and paired G-buffers, training separate inverse and forward rendering models. UniRelight~\cite{he2025unirelight} further extends this to end-to-end video relighting. However, both methods input environment maps directly without explicit light transport guidance, limiting generalization in complex scenes with occlusions. In contrast, our method targets general dynamic scenes beyond portraits and employs physically-based irradiance cues to achieve more accurate lighting control.

Several related systems target narrower capture settings or different control interfaces. SwitchLight~\cite{kim2024switchlight} decomposes a single portrait using a physics-guided architecture and relights it from an environment map, whereas DifFRelight~\cite{he2024diffrelight} learns subject-specific facial-performance relighting from light-stage data. Relightable Full-Body Gaussian Codec Avatars~\cite{wang2025fullbody} models local and non-local light transport for captured, personalized avatars. ReLumix~\cite{wang2025relumix} propagates the illumination of one artist-relit reference frame through a video rather than taking a target environment map directly. The commercial Runway Relight Scene system~\cite{runwayrelightscene} instead uses text descriptions to edit the atmosphere of short videos. These methods demonstrate complementary progress, but do not address pixel-aligned, environment-map-controlled relighting of general monocular human-centric videos in the wild.

RGB$\leftrightarrow$X~\cite{zeng2024rgb} shows pixel-aligned irradiance maps can serve as lighting control signals, but require an existing clean irradiance map as input. DiLightNet~\cite{zeng2024dilightnet}, PractiLight~\cite{erel2025practilight}, and Lightit~\cite{kocsis2024lightit} compute irradiance cues from target lighting and coarse depth, but misalignment with the input image due to noisy geometry undermines relighting quality. Portrait-specific relighting methods such as Total Relighting~\cite{pandey2021total} focus on human faces and use image-based lighting without path-traced shadows, which is suitable for portrait photography but less applicable to general scenes.

\paragraph{Joint cross-modal modeling.}
Diffusion models have been applied to jointly model pixel-aligned attributes such as depth~\cite{li2025simple,byung2025jointdit,xi2025omnivdiff,chen20254dnex}, motion~\cite{chefer2025videojam}, and PBR attributes~\cite{xi2025ctrlvdiff,he2025unirelight}. Joint modeling approaches include dual-stream cross-attention~\cite{li2025simple,byung2025jointdit} and various single-model injection paradigms~\cite{chefer2025videojam,xi2025omnivdiff,xi2025ctrlvdiff,he2025unirelight,chen20254dnex}. Most such methods assume clean auxiliary modalities; in contrast, we must handle imperfect irradiance cues with substantial geometry artifacts—a domain-gap issue largely unexplored in prior work. The use of per-modality asynchronous timestep schedules has been explored in UniDiffuser~\cite{bao2023one}, AVDiT~\cite{wang2024av}, and Uni-Renderer~\cite{chen2025uni}, but none addresses imperfect auxiliary modalities at inference time. Our key contribution is applying this idea to the RGB-irradiance pairing, combined with domain matching and hijacked inference to handle noisy real-world irradiance cues.

\paragraph{Multi-illumination dataset.}
Training a video relighting network requires paired multi-illumination data with lighting annotation. Existing datasets~\cite{murmann2019dataset,teufel2025humanolat,liu2023openillumination} mainly cover static images in restricted scenarios, and prior works~\cite{liang2025diffusion,he2025unirelight} use only simple rendered scenes. Unreal Engine 5~\cite{unrealengine} enables high-quality human-centric multi-illumination video synthesis with accurate lighting annotation.

\begin{figure}[t]
    \centering
    \includegraphics[width=\linewidth]{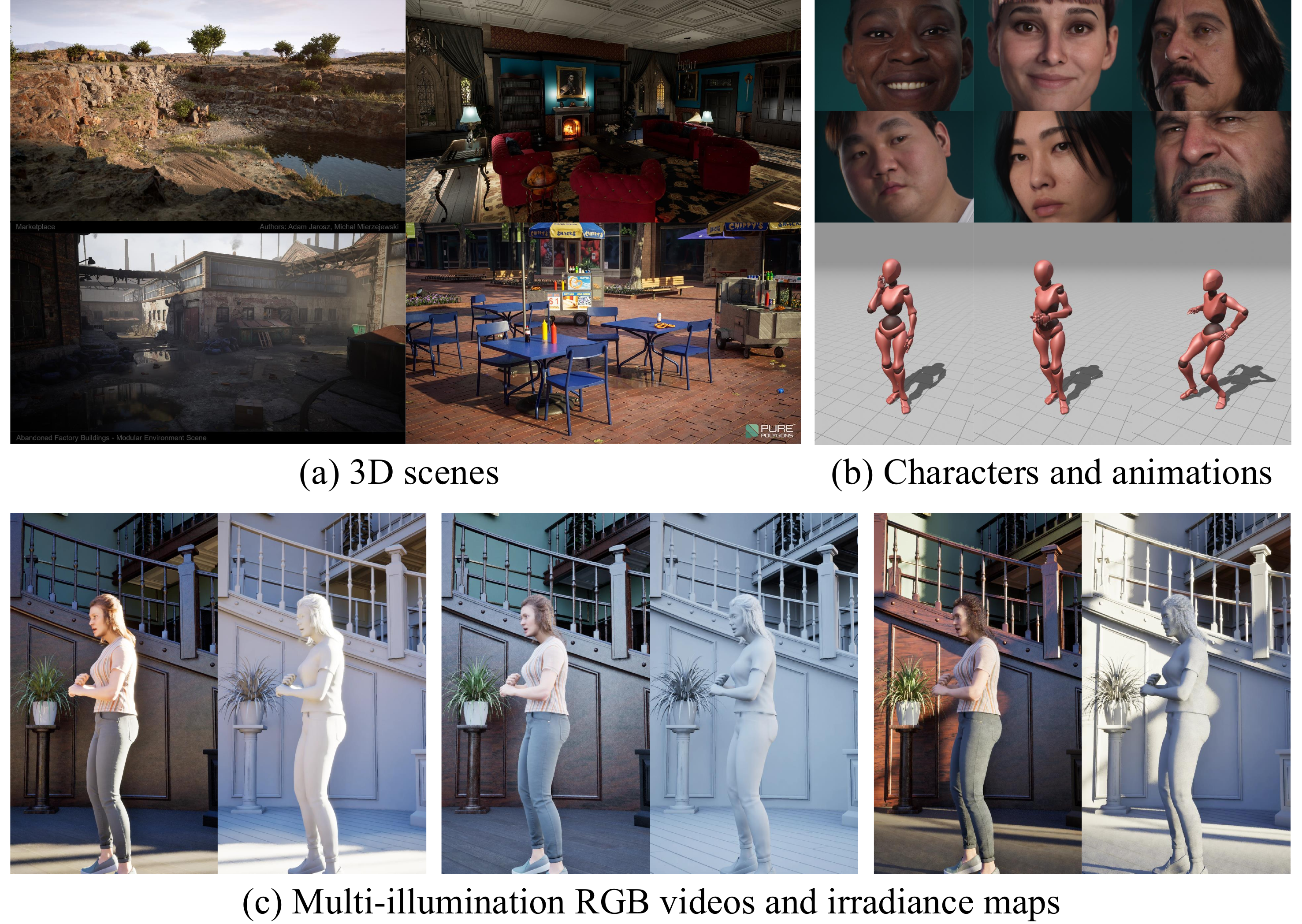}
    \caption{Illustration of our introduced multi-illumination synthetic dataset.
   }
    \label{fig::dataset}
\end{figure}

\section{Preliminary}
Our study builds on a transformer-based latent diffusion model~\cite{wan2025}. The architecture comprises a 3D Variational Autoencoder (VAE) and a diffusion transformer (DiT) of spatio-temporal transformer blocks. Each block integrates 3D spatio-temporal self-attention, cross-attention, feed-forward network, and gate modulation. The generative process can be defined as a mapping between samples $x_1$ from a noise distribution to samples from a data distribution $x_0$ in terms of an ordinary differential equation (ODE),
\begin{align}
dz_t=v_{\Theta}(z_t,t)dt,
\end{align}
where the velocity $v$ is parameterized by the weights $\Theta$ of the diffusion transformer and $t$ denotes the timestep.
The noise scheduling follows the Rectified Flow framework~\cite{liu2022flow, esser2024scaling}. Let $z_0$ denote a clean latent, $\epsilon\sim\mathcal{N}(0,I)$ a Gaussian-noise latent, and $t\in[0,1]$ a continuous noise level, where $t=0$ is clean data and $t=1$ is noise. Rectified Flow defines a straight interpolation
\begin{align}
z_t=\psi_t(z_0|\epsilon)= (1-t)z_0+t\epsilon.
\end{align}
The target velocity along this path is $u_t=\epsilon-z_0$. The Conditional Flow Matching objective regresses this vector field with an MSE loss:
 \begin{align}
    L_{\mathrm{CFM}}=\mathbb{E}_{t,z_0,\epsilon}\left[\lVert v_\Theta(z_t,t)-(\epsilon-z_0)\rVert^2_2\right],
\end{align}
where $v_\Theta$ is the predicted velocity. At inference, integrating the learned ODE from noise toward data produces a clean sample.

\section{Training Dataset}\label{sec::synth_dataset}
We build an automated UE5 pipeline for multi-illumination videos. Our assets include 30 3D environments, 60 characters, 600 body animations, 20 facial expressions, and 821 HDR environment maps. Figure~\ref{fig::dataset} overviews the dataset. For each scene, a character is randomly placed and animated, while the camera is assigned random distance, rotation, and oscillation. Ray intersection tests prevent interpenetration. We use UE5's real-time global illumination system Lumen~\cite{Tan2024} for efficient low-frequency approximation, and add a directional light to match the brightest region of the environment map. LightingOnly mode renders the irradiance map synchronously. Each scene is rendered under 10 illuminations, with 20\% using rotary lighting. In total, we render 68K videos at  $832\times480$ resolution, 94 frames each, with paired irradiance maps. Lighting parameters are recorded, and prompts are generated using a video understanding model~\cite{damonlpsg2025videollama3}. Among the 30 scene environments, 12 are indoor scenes and 18 are outdoor scenes. The 821 HDR environment maps are collected from public resources such as Poly Haven, covering diverse illumination conditions including outdoor daytime, indoor, and outdoor nighttime settings.  We pair the MetaHuman characters with Mixamo animations and mainly keep in-place motions such as talking, dancing, and fighting, while filtering out large-displacement motions such as walking and running to better match our camera setup.

The set contains 31 male-presenting and 29 female-presenting characters; 22 dark-, 12 medium-, and 26 light-skinned characters; and 4 child, 43 adult, and 13 older characters.
Our dataset focuses on human-centric indoor and outdoor scenarios, enabling precise lighting control over human subjects. Despite being trained purely on synthetic data, our model generalizes well to real-world videos. We attribute this to the high photorealism of UE5-rendered content, which substantially reduces the sim-to-real gap compared to prior Blender-based synthetic datasets. This aligns with observations from prior works trained exclusively on UE-rendered videos, such as ReCamMaster~\cite{bai2025recammaster} and Context-as-Memory~\cite{yu2025context}.

\section{Method}
We first describe how to construct the irradiance cue $\mathbf{E}^{\text{cue}}$ for an input video $\mathbf{V}^I$, then present our asynchronous RGB-irradiance joint modeling and hijacked inference strategy for generating the relit video $\mathbf{V}$ and refined irradiance map $\hat{\mathbf{E}}^{\text{cue}}$.

\subsection{Irradiance cue}\label{sec::irradiance_cue}
An irradiance cue describes diffuse light transport obtained by rendering scene geometry with target lighting. Previous works~\cite{zeng2024dilightnet,erel2025practilight} use monocular depth~\cite{bhat2023zoedepth} and unproject it to a mesh. We instead use  monocular point-cloud estimation~\cite{wang2025moge,wang2025moge2} to obtain metric-level geometry and smoother meshes.  We convert each estimated point cloud into a per-frame mesh and remove triangles spanning depth discontinuities with a fixed threshold-based mask. We assign a spatially uniform white diffuse BRDF , illuminate the mesh with the target environment map, and render it using Mitsuba 3~\cite{Mitsuba3}. Repeating this process independently for every frame yields the irradiance cue $\mathbf{E}^{\text{cue}}\in\mathbb{R}^{F\times H\times W \times 3}$ for an input video $\mathbf{V}$.  Per-frame estimation avoids accumulating global reconstruction drift, but geometry errors,  removed boundary triangles, and temporal inconsistency still create a substantial gap from the ground-truth irradiance $\mathbf{E}$ (Figure~\ref{fig::infer}).

\subsection{Joint modeling with separate timestep scheduling}
\paragraph{Noise Corruption.} Directly using the imperfect irradiance cue $\mathbf{E}^{\text{cue}}$ causes model confusion due to pixel misalignment. To retain its low-frequency lighting prior while suppressing artifacts, we corrupt it with Gaussian noise $\epsilon$:
\begin{align}\label{equation::noise_irradiance_cue}
    \mathbf{E}^{\text{cue}}_t = (1-t)\mathbf{E}^{\text{cue}}+t\epsilon.
\end{align}
In this way, the high-frequency artifacts are removed and the global light transport prior is preserved. Simple hole filling or spatial blurring cannot reliably reconstruct light transport around undefined geometry and may erase useful low-frequency lighting structure; joint denoising instead learns how much of the corrupted cue to retain. To exploit this prior, we jointly model irradiance maps and RGB videos within the diffusion framework.

\paragraph{Model design.} Given an input video $\mathbf{V}^I$ and a text prompt $p$, we train a model $v_\Theta$ that jointly denoises the relit RGB video $\mathbf{V}$ and irradiance map $\mathbf{E}$ using separate timesteps $t_V$ and $t_E$. This allows the model to learn denoising under mismatched noise levels and enables inference with noisy irradiance conditions (Section~\ref{sec::inference}).
As illustrated in Figure~\ref{fig::pipeline}, we concatenate the latents of noisy RGB video $\mathbf{z}_{t_V}^V$, noisy irradiance map $\mathbf{z}_{t_E}^E$ and input video $\mathbf{z}_0^I$ along the frame dimension and feed into the transformer. This simple strategy allows the 3D spatial-temporal self-attention to cover input, irradiance and relit frames, enabling cross-modal understanding. While asynchronous timestep scheduling has been explored in prior multi-modal diffusion work~\cite{bao2023one,wang2024av,chen2025uni}, our contribution lies in applying it to the RGB-irradiance pairing, where the auxiliary modality is imperfect at inference time. This motivates our domain matching strategy and hijacked inference pipeline, which are essential for bridging the gap to noisy real-world irradiance cues.

\paragraph{Modality disambiguation.} Since the three input modalities share pixel alignment, we apply the same RoPE~\cite{su2024roformer} to assign shared spatial-temporal positions. To distinguish modalities, we introduce a learnable modality embedding on top of the timestep embedding, whose gate modulation already conditions features on timestep. Unlike prior designs that inject modality embeddings into latent features, this better avoids modality mixture in our setting.

\paragraph{Domain matching} Since the model only observes the ground-truth irradiance map $\mathbf{E}$ during standard training, conditioning on imperfect irradiance cues at inference introduces a domain gap. We address this with a domain matching paradigm: at random training iterations, we replace the noisy irradiance map $\mathbf{z}_t^E$ with the noisy cue $\mathbf{z}_t^{E^{\text{cue}}}$. This encourages denoising towards a probability path closer to the high-quality manifold and improves both RGB and irradiance generation. This design is fundamentally different from previous joint cross-modal modeling methods, which usually assume clean and reliable modalities during both training and inference. In contrast, our framework is explicitly designed for imperfect irradiance cues with domain discrepancies, which are common in practical relighting settings. It also differs from prior irradiance-conditioned methods that directly feed the estimated cue into the generator, making them more vulnerable to accumulated geometry and rendering errors.

\subsection{Hijacked asynchronous inference strategy}\label{sec::inference}
In Section~\ref{sec::irradiance_cue} we show how to construct the irradiance cue $\mathbf{E}^{\text{cue}}$, given an input video and a target lighting condition. In order to utilize it during inference, we design a hijacked conditional inference strategy that guides the sampling process towards the probability path indicated by the irradiance cue. We illustrate this process in Figure~\ref{fig::infer}. Specifically, we obtain the noisy irradiance cue $\mathbf{E}^{\text{cue}}_{t_h}$ with Eq.~\ref{equation::noise_irradiance_cue}, where $t_h$ is the hijack timestep. Before $t_h$, at each timestep only the noisy RGB latents are updated while the irradiance latents remain fixed at $\mathbf{E}^{\text{cue}}_{t_h}$. After $t_h$, the two modalities are denoised together. This strategy effectively utilizes low-frequency lighting prior provided by the noisy irradiance map, while removing high-frequency artifacts by joint denoising. The hyperparameter $t_h$ controls the amount of noise that is added to the irradiance cue, offering a flexible tradeoff between generation quality and lighting fidelity. In practice, we find that $t_h=0.6T$ works well for most cases. When the irradiance cue is reliable, a smaller $t_h$ helps preserve more high-frequency appearance details. When the cue is noisy or less aligned, a larger $t_h$ is preferable because it suppresses artifacts more aggressively.

\begin{figure*}[ht]
    \centering
    \includegraphics[width=\linewidth]{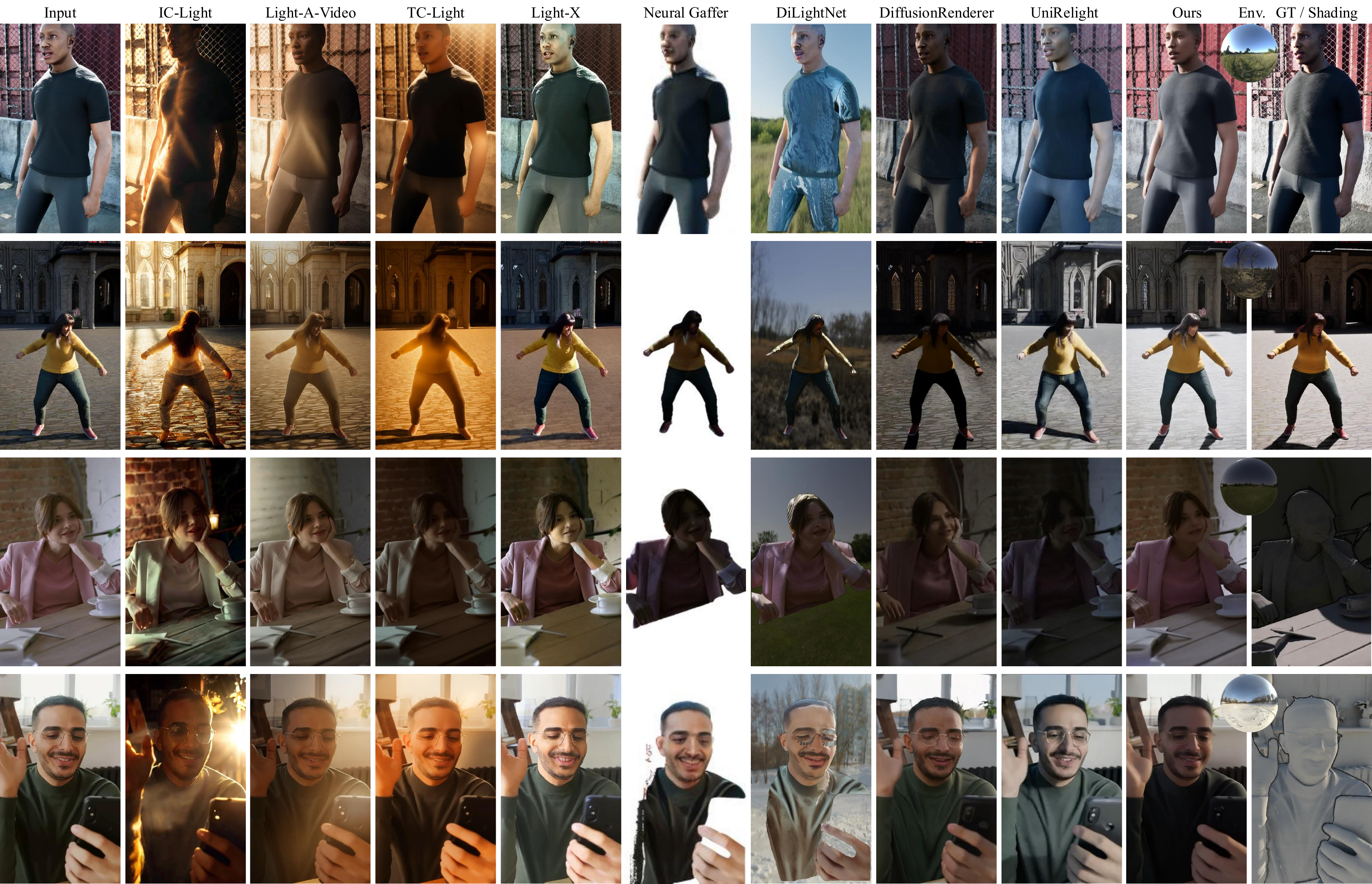}
    \caption{
    Qualitative comparison on vertical \textit{SynthStatic} and \textit{HumanVid} videos. The last row shows synthetic ground truth and real-world irradiance cues.}
    \label{fig::comparison_vertical}
\end{figure*}

\begin{figure*}[ht]
    \centering
    \includegraphics[width=\linewidth]{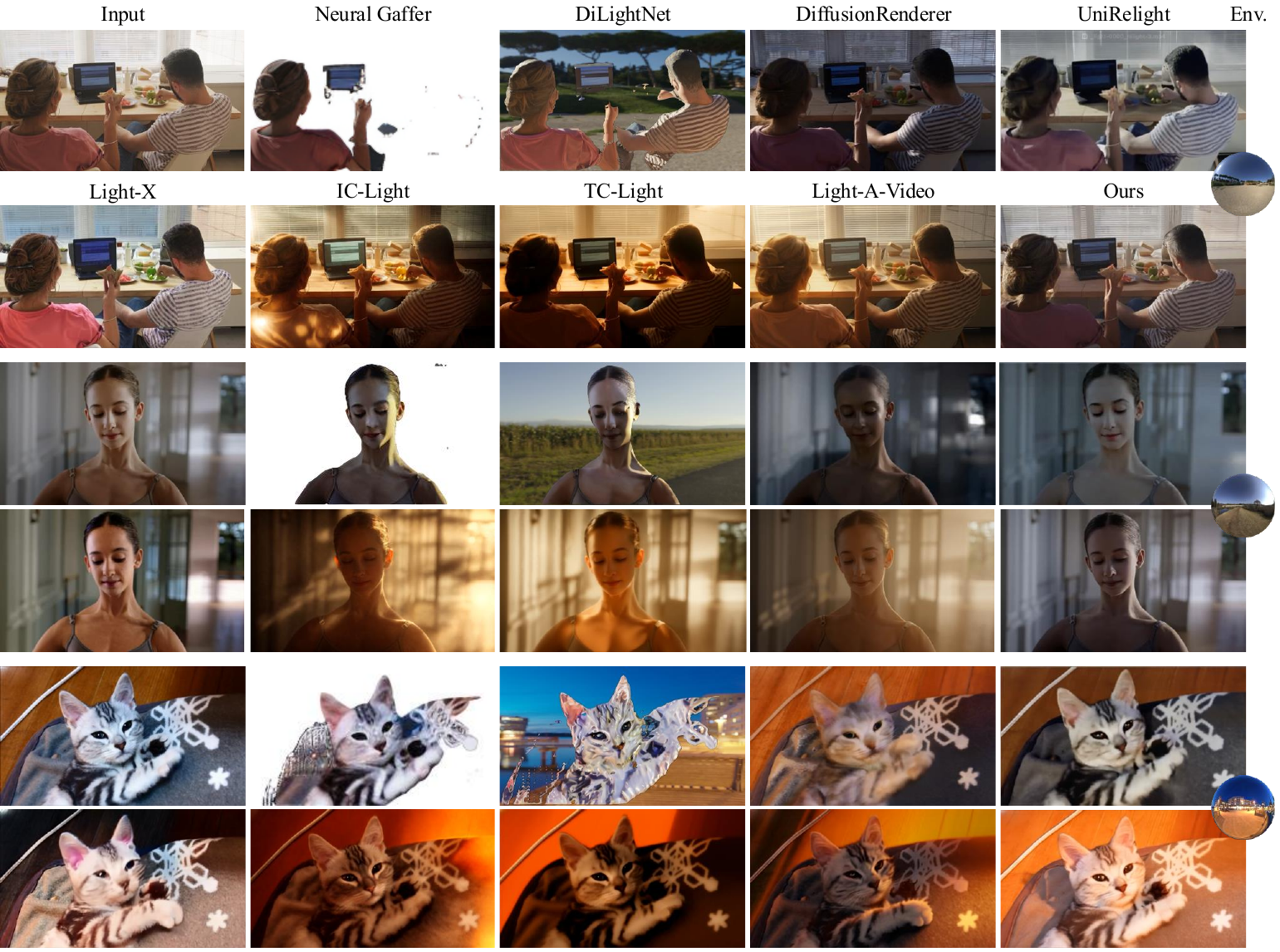}
    \caption{ Qualitative comparison on horizontal images.
   }
    \label{fig::comparison_horizontal}
\end{figure*}

\section{Experiments}
\paragraph{Implementation details.} We fine-tune Wan2.2-TI2V-5B~\cite{wan2025} for 12K steps at  $832\times480$ resolution using a learning rate of  $10^{-5}$ and batch size 16. Domain matching is applied only during the last 4K steps. We apply this second-stage training after the first-stage joint modeling has largely converged, and empirically find that 4K steps are sufficient for the model to adapt to the imperfect irradiance-cue distribution. To further mitigate the visual characteristics of synthetic training data, we adopt a noise augmentation strategy inspired by ReCamMaster~\cite{bai2025recammaster}: a moderate level of noise (equivalent to 200--500 diffusion timesteps) is added to the conditional video latent during training, which discourages the model from relying on domain-specific synthetic cues and improves robustness at real-world inference time. We set the hijack timestep to $t_h=0.6T$ for every experiment and do not retune it case by case.

\paragraph{Baselines} We compare our method with the publicly available video relighting method DiffusionRenderer~\cite{liang2025diffusion}, UniRelight~\cite{he2025unirelight} and Light-X~\cite{liu2025light}. Image relighting methods DiLightNet~\cite{zeng2024dilightnet} and Neural-Gaffer~\cite{jin2024neural} are also included. We adopt the Cosmos~\cite{agarwal2025cosmos} implementation version of DiffusionRenderer for a fair comparison. These methods allow environment maps as control signals. We additionally include prompt-based relighting methods, including IC-Light~\cite{zhang2025scaling}, Light-A-Video~\cite{zhou2025light}, TC-Light~\cite{liu2025tc}. For environment-map-conditioned methods we use the identical target HDR map; prompt-based methods instead receive best-effort captions derived from the target lighting, so exact direction and intensity cannot be guaranteed.

\paragraph{Datasets} We use the test subset of our synthetic dataset for quantitative evaluation. It contains 100 test cases: 50 static-lighting cases (\textit{SynthStatic}) and 50 rotary-lighting cases (\textit{SynthDynamic}), each with an input video, relit target, and target environment map. For qualitative comparison, test cases are sampled from the HumanVid~\cite{wang2024humanvid} dataset. Since there is no publicly available video relighting dataset for quantitative evaluation, we follow  previous video relighting methods such as DiffusionRenderer~\cite{liang2025diffusion} and UniRelight~\cite{he2025unirelight} by using our proposed synthetic dataset for  the video benchmark. To test cross-dataset generalization independently of our synthetic renderer, we additionally evaluate static images from HumanOLAT~\cite{teufel2025humanolat}.
\paragraph{Metrics.} We report PSNR, SSIM, and LPIPS~\cite{zhang2018unreasonable}. We also conduct a user study on visual quality and illumination accuracy. Following prior works~\cite{jin2024neural, liang2025diffusion, he2025unirelight}, we avoid per-channel rescaling and instead compare results directly with ground truth, enabling faithful evaluation of relighting fidelity.

\begin{table*}[t]
\centering
\caption{Quantitative comparison on  \textit{SynthStatic} and \textit{SynthDynamic}, and user-study scores. VQ and LF denote visual quality and lighting fidelity. Best and second-best results are bold and underlined, respectively.}
\label{tab:comparison}
\begin{tabular}{l | c c c | c c c | c c }
\toprule
 & \multicolumn{3}{c|}{\textit{SynthStatic}}
 & \multicolumn{3}{c|}{\textit{SynthDynamic}}
 & \multicolumn{2}{c}{User Study} \\
 & PSNR $\uparrow$ & SSIM $\uparrow$ & LPIPS $\downarrow$
 & PSNR $\uparrow$ & SSIM $\uparrow$ & LPIPS $\downarrow$
 &  VQ $\uparrow$ & LF $\uparrow$ \\
\midrule
IC-Light        & 12.79 & 0.508 & 0.329 & 12.76 & 0.512 & 0.317 & 3.28 & 2.10 \\
Light-A-Video        & 12.18 & 0.390 & 0.401 & 11.82 & 0.375 & 0.440 & 3.41 & 3.01 \\
TC-Light        & 10.74 & 0.433 & 0.472 & 9.87 & 0.366 & 0.539 & 2.73 & 2.45 \\
\midrule
DiLightNet           & 11.76 & 0.507 & 0.336 &  11.21 & 0.493 & 0.371 & 1.19 & 1.31 \\
Neural Gaffer        & 13.72 &  0.629 & \best{0.269} & 12.25 & 0.607 &  0.327 & 2.40 & 2.57 \\
Light-X        & 14.03 & 0.582 & 0.285 & 12.40 & \best{0.619} & 0.308 & 3.51 & 2.69 \\
DiffusionRenderer (Cosmos) & 15.27 & 0.615 & 0.306 & 13.48 & 0.589 & 0.345 & 3.15 & 3.04 \\
UniRelight & \best{15.83} & \bestbold{0.661} & 0.288 & \best{14.90} & 0.602 & \best{0.303} & \best{3.92} & \best{3.82} \\
\midrule
Ours                 & \bestbold{16.62} & \best{0.656} & \bestbold{0.251} & \bestbold{16.39} & \bestbold{0.631} & \bestbold{0.263} & \bestbold{3.98} & \bestbold{4.53} \\
\bottomrule
\end{tabular}
\end{table*}

\subsection{Evaluation}

\paragraph{Qualitative comparison} Figures~\ref{fig::comparison_vertical} and ~\ref{fig::comparison_horizontal} compare vertical and horizontal cases, respectively; the latter includes \textit{SynthStatic} examples. Prompt-based methods including IC-Light, TC-Light, and Light-A-Video produce appealing lighting styles, but cannot precisely control lighting direction or ambient color because text prompts are ambiguous. These methods often produce warm tones and can retain hard input shadows, as in the second case of Figure~\ref{fig::comparison_vertical}. Light-X and Neural Gaffer inherits much of the input lighting distribution and mainly modifies color tone, while DiLightNet produces low-quality results caused by inaccurate material estimates. DiffusionRenderer and UniRelight performs reasonably well in terms of light direction and color. However, inaccurate material estimation yields unrealistic skin and clothing textures (Figs.~\ref{fig::comparison_horizontal} and~\ref{fig::comparison_vertical}). It also causes identity drift, demonstrated on the third case of Figure~\ref{fig::comparison_horizontal}. More importantly, implicit light transport modeling sometimes fails to produce physically correct results. In Fig.~\ref{fig::comparison_vertical} (third case), DiffusionRenderer mistakes yellow ambient light for the main source, ignoring the actual moon light.  In contrast, our method models light transport with guidance from the path-traced irradiance cue, facilitating physically accurate illumination control. Although trained on human-centric data, our method also generalizes to the cat example in Figure~ \ref{fig::comparison_horizontal}.

\paragraph{Quantitative comparison} Table 1 confirms the qualitative observation that DiffusionRenderer and UniRelight respond more faithfully to environment maps than the other baselines. We evaluate on 100 test cases in total, with 50 cases from \textit{SynthStatic} and 50 from \textit{SynthDynamic}. For prompt-based methods, we ask GPT 5.4 to describe the lighting in the ground truth video and use the caption as text prompt input. For image-based methods, we perform relighting frame by frame.
Our method  obtains the best PSNR and LPIPS on \textit{SynthStatic}, while UniRelight obtains the best SSIM. On \textit{SynthDynamic}, our method is best on all three metrics. The larger advantage under rotating illumination indicates that explicit irradiance guidance improves controllability when the target lighting changes over time.

\begin{figure*}[ht]
    \centering
    \includegraphics[width=0.9\linewidth]{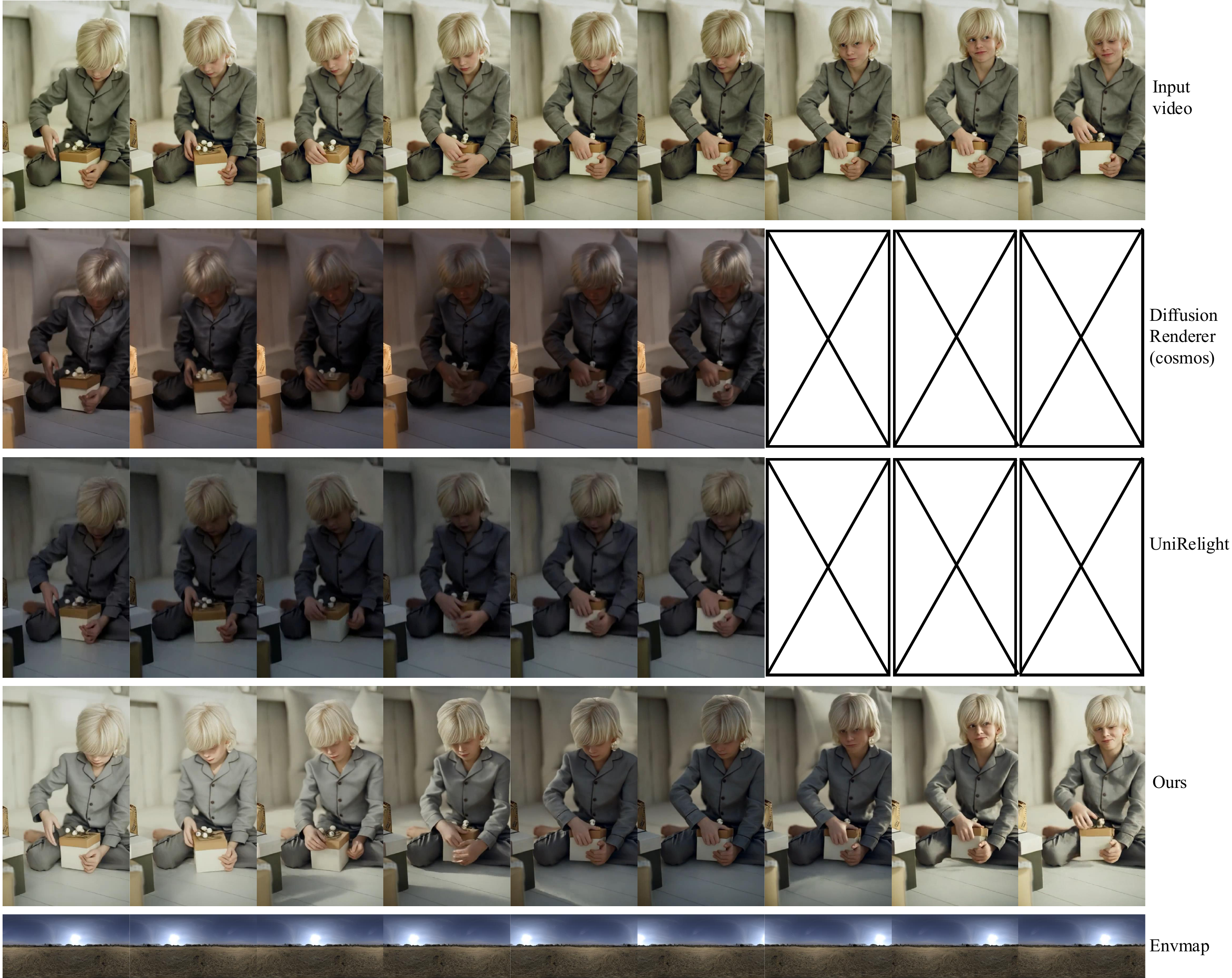}
    \caption{Results under dynamic lighting. Images are sampled every 10 frames while the environment map  rotates through $360^\circ$.
   }
    \label{fig::rotary_light}
\end{figure*}

\begin{figure}[ht]
    \centering
    \includegraphics[width=0.98\linewidth]{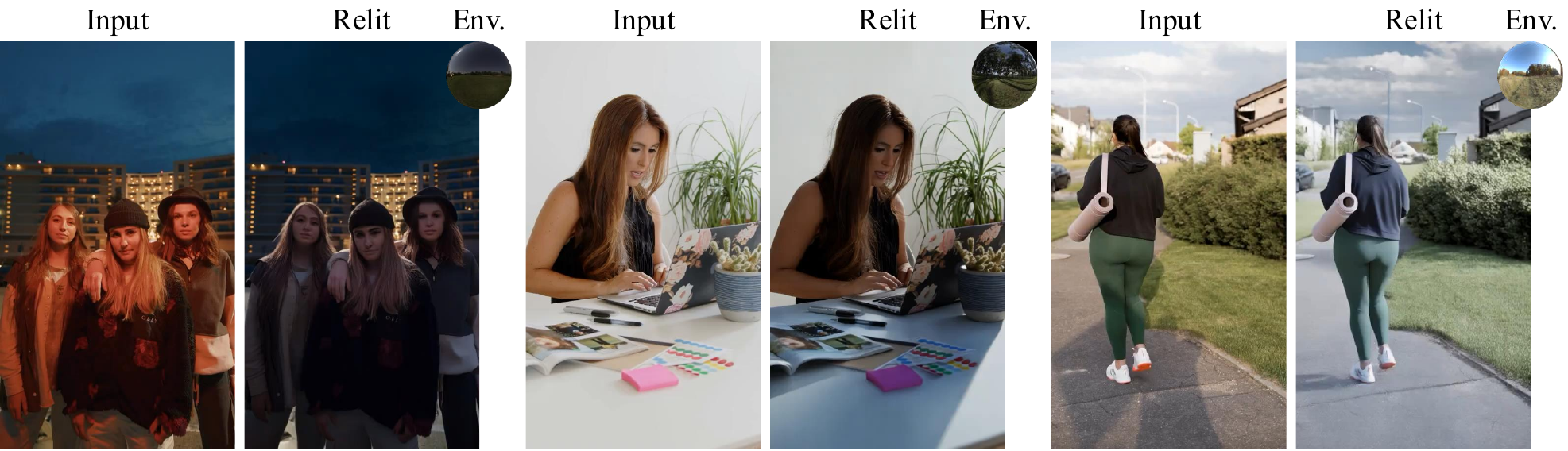}
    \caption{More results on vertical images from indoor and outdoor scenes. In the second example, the light source appears in the upper-right region of the chrome ball. We  show a black background for clarity.
   }
    \label{fig::more_vertical}
\end{figure}

\begin{figure*}[ht]
    \centering
    \includegraphics[width=0.98\linewidth]{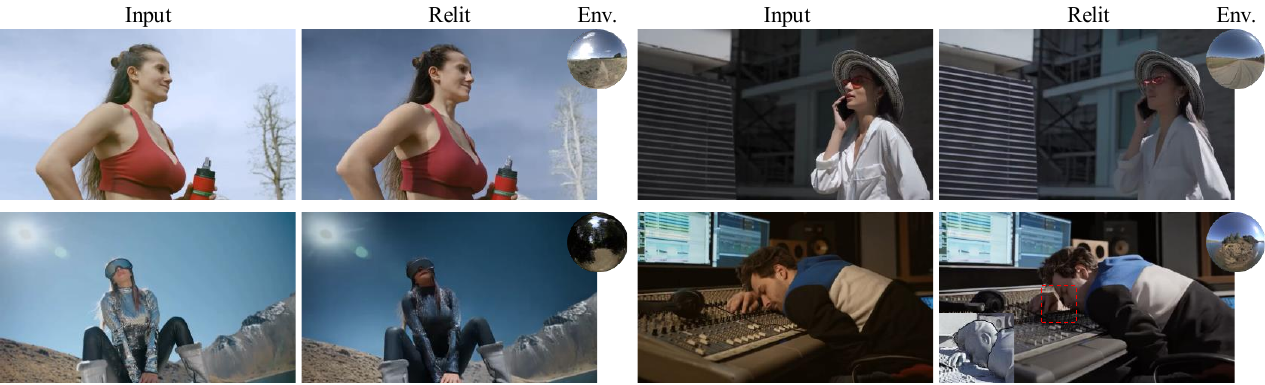}
    \caption{More results and failure cases on horizontal images from indoor and outdoor scenes. The bottom-left example shows the limitation on highly reflective surfaces;  the bottom-right example shows light leakage around the hand caused by inaccurate depth near a discontinuity.
   }
    \label{fig::more_horizontal}
\end{figure*}

\begin{figure}[ht]
    \centering
    \includegraphics[width=\linewidth]{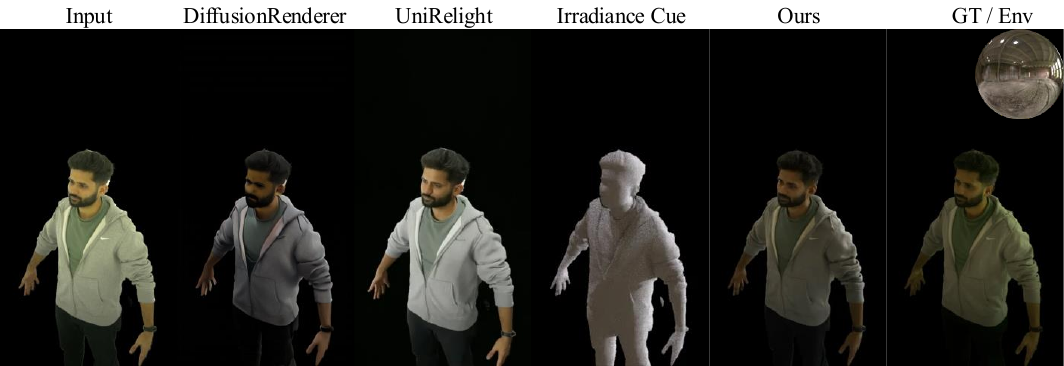}
    \caption{Qualitative comparison on HumanOLAT dataset.}
    \label{fig::humanolat}
\end{figure}

\paragraph{HumanOLAT evaluation}
We synthesize ground-truth relit images from HumanOLAT OLAT captures under 100 HDR environment maps. We apply the provided foreground masks before computing metrics and do not perform per-channel rescaling. Table~\ref{tab:humanolat} shows that our model, trained without HumanOLAT data, achieves the best performance among all evaluated methods. Qualitative comparison is presented in Figure~\ref{fig::humanolat}

\begin{table}[t]
\centering
\caption{Cross-dataset evaluation on HumanOLAT.}
\label{tab:humanolat}
\begin{tabular}{lccc}
\toprule
Method & PSNR$\uparrow$ & SSIM$\uparrow$ & LPIPS$\downarrow$ \\
\midrule
Neural Gaffer & 17.10 & 0.881 & 0.151 \\
DiLightNet & 15.68 & 0.847 & 0.142 \\
Light-X & 18.01 & 0.862 & 0.161 \\
DiffusionRenderer & \best{22.32} & \best{0.912} & \best{0.114} \\
UniRelight & 19.54 & 0.871 & 0.136 \\
Ours & \bestbold{23.89} & \bestbold{0.928} & \bestbold{0.078} \\
\bottomrule
\end{tabular}
\end{table}

\begin{figure}[ht]
    \centering
    \includegraphics[width=\linewidth]{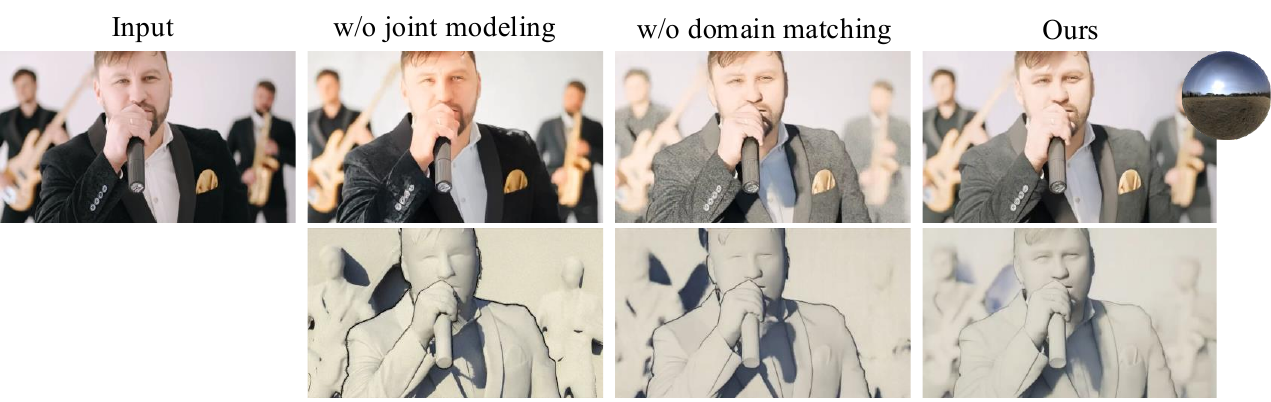}
    \caption{
    Qualitative ablation. From left: input, direct irradiance conditioning without joint modeling, without domain matching, and full model.}
    \label{fig::ablation}
\end{figure}

\paragraph{User study} A user study evaluated perceptual quality on in-the-wild videos. Participants viewed the relighting results, the input video, and the environment map, and rated methods from 1 to 5 on visual quality (VQ) and lighting fidelity (LF). The study involved 13 in-lab users, and average scores are reported in Table~\ref{tab:comparison}. We evaluate 15 indoor and outdoor test cases in total. Before the study, we explain that visual quality measures the overall perceptual realism, including artifacts and temporal stability, while lighting fidelity evaluates how well the generated result matches the target illumination in terms of light direction, shadow, and highlight consistency. Compared to baselines, our results are consistently preferred on both dimensions.

\paragraph{Dynamic scenes with dynamic illumination} Thanks to explicit light transport guidance from asynchronous RGB-irradiance joint modeling, our method naturally extends to dynamic scenes with dynamic illumination. Figure~\ref{fig::rotary_light} shows results sampled every 10 frames. Our method faithfully follows the sun direction and produces physically plausible lighting effects.

\paragraph{Runtime}
For an 81-frame, $832\times480$ video sampled with 50 diffusion steps on one NVIDIA A800 GPU, the full pipeline takes 223\,s: 22.8\,s for depth estimation, 38.2\,s for mesh conversion, 18.8\,s for Mitsuba rendering, and 141.9\,s for neural generation (minor overhead accounts for the remaining time). Under the same evaluation setting, UniRelight, DiffusionRenderer, and Light-A-Video take 1077\,s, 1629\,s, and 1822\,s, respectively. Thus, external reconstruction and rendering are not the main runtime bottleneck.

\section{Ablation study}~\label{sec::ablation}
\paragraph{Qualitative ablation} We conduct ablation study on our asynchronous RGB-irradiance joint modeling and domain matching strategy with qualitative results in Figure~\ref{fig::ablation}. Without the asynchronous joint modeling training paradigm, simply using the irradiance cue as control signal, the model fails to follow the light transport. Joint modeling alone does not sufficiently bridge the domain gap; domain matching is essential for producing pixel-aligned irradiance maps and physically plausible RGB results. Without it, the denoised irradiance maps exhibit black edges and blurred details, leading to suboptimal outputs. In contrast, the full recipe generates pixel-aligned RGB and irradiance images, producing physically accurate and visually plausible results.

\begin{table}[t]
\centering
\caption{Quantitative ablation on major components. }
\label{tab:ablation_components}
\begin{tabular}{lccc}
\toprule
Variant & PSNR$\uparrow$ & SSIM$\uparrow$ & LPIPS$\downarrow$ \\
\midrule
HDR map conditioning & 13.79 & 0.585 & 0.331 \\
irradiance conditioning & 14.10 & 0.578 & 0.327 \\
+ Joint modeling & 14.06 & 0.572 & 0.321 \\
+ Hijacked conditional generation & 14.69 & 0.582 & 0.306 \\
+ Asynchronous timestep scheduling & 15.20 & 0.611 & 0.298 \\
+ Domain matching strategy (Ours) & \textbf{15.69} & \textbf{0.621} & \textbf{0.282} \\
\bottomrule
\end{tabular}
\end{table}

\begin{table}[t]
\centering
\caption{Ablation on modality embedding designs.}
\label{tab:ablation_embedding}
\begin{tabular}{lccc}
\toprule
Variant & PSNR$\uparrow$ & SSIM$\uparrow$ & LPIPS$\downarrow$ \\
\midrule
Add to latent & 15.38 & 0.615 & 0.299 \\
Concat to latent & 15.42 & 0.620 & 0.305 \\
Add to time embedding (Ours) & \textbf{15.69} & \textbf{0.621} & \textbf{0.282} \\
\bottomrule
\end{tabular}
\end{table}

\paragraph{Quantitative ablation} We report quantitative ablation results in Tables~\ref{tab:ablation_components} and \ref{tab:ablation_embedding}.
We additionally train a no-irradiance variant that receives only the target HDR environment map under the same data and optimization protocol, analogous to HDR-conditioned implicit relighting.
All ablation models are trained for 6K steps, with domain matching applied during the final 2K steps. HDR map conditioning does not provide pixel-aligned guidance for light distribution, which produces less accurate relighting results compared to irradiance conditioning. Joint modeling irradiance and RGB alone brings limited gains because the imperfect irradiance cue still causes a noticeable train-test mismatch. Introducing hijacked conditional generation and asynchronous timestep scheduling progressively improves both lighting fidelity and visual quality, while the domain matching strategy delivers the final performance boost by explicitly aligning the cue distribution with inference-time inputs. We also compare different modality embedding strategies and find that injecting the modality cue through the time embedding performs best, likely because it modulates the denoising process globally and reduces the ambiguity between RGB appearance and irradiance guidance.

\section{Discussion}
We present JointLight, a human-centric video relighting framework that jointly models RGB and irradiance asynchronously. We also introduce a high-quality human-centric multi-illumination synthetic dataset covering indoor and outdoor scenarios, paired with ground-truth irradiance maps. Trained exclusively on this dataset, our model achieves state-of-the-art relighting quality and fidelity.

\paragraph{Robustness to geometry errors}
Our method depends on monocular geometry only through the rendered irradiance cue. Because geometry is estimated independently per frame, errors do not accumulate as global reconstruction drift. Noise injection suppresses high-frequency defects such as jagged boundaries, holes, and aliased shadows, but cannot recover light transport that is absent from severely incorrect geometry. Aggressive subject or camera motion can also produce blur and occlusion errors. Incorrect depths or holes near discontinuities may distort the path-traced cue and can cause light leakage (the bottom-right case in Figure~\ref{fig::more_horizontal}).

\paragraph{Limitations and Future Work}
First, irradiance provides diffuse transport only and therefore cannot explicitly or physically control full specular transport. Video priors can synthesize plausible moderate highlights on skin and hair, but these effects are not guaranteed to follow the target illumination; polished metal, mirrors, and glass remain difficult. A failure case is shown in the bottom-right case of Figure~\ref{fig::more_horizontal}. Second, suppressing high-frequency cue errors can also attenuate texture-rich appearance, occasionally over-smoothing fine facial details around the eyes and mouth. Future work could jointly reason about geometry, material, diffuse and specular transport, and use detail-preserving or preference-based post-training. Extending the control representation beyond environment maps to local spotlights or flashlights is another promising direction.

\begin{acks}
This work was supported by the National Natural Science Foundation of China (No. 62322210, 62561160115), the Beijing Major Science and Technology Project (No. Z251100008125009), the Royal Society (IEC\textbackslash NSFC\textbackslash 233596), and the Innovation Funding of ICT, CAS (No. E561100).
\end{acks}

\bibliographystyle{ACM-Reference-Format}
\bibliography{references}

\appendix